\documentclass[conference,a4paper]{APSIPA2021}
\usepackage{multirow}
\usepackage[dvips]{graphicx}
\usepackage{amsmath}
\usepackage[psamsfonts]{amssymb}
\usepackage{amsxtra}
\usepackage{makecell}
\usepackage{threeparttable}

\begin{document}

\title{OLR 2021 CHALLENGE: DATASETS, RULES AND BASELINES}

\author{
\authorblockN{%
Binling Wang\authorrefmark{2}$^a$,
Wenxuan Hu\authorrefmark{2}$^a$, 
Jing Li\authorrefmark{2},
Yiming Zhi\authorrefmark{2},
Zheng Li\authorrefmark{3},
\\
Qingyang Hong\authorrefmark{2}$^*$,
Lin Li\authorrefmark{3}$^*$,
Dong Wang\authorrefmark{4}\authorrefmark{5},
Liming Song\authorrefmark{6} and
Cheng Yang\authorrefmark{6}
}
\authorblockA{%
\authorrefmark{2}
School of Informatics, Xiamen University\\
\authorrefmark{3}
School of Electronic Science and Engineering, Xiamen University\\
\authorrefmark{4}
Center for Speech and Language Technologies, Tsinghua University\\
\authorrefmark{5}
Beijing National Research Center for Information Science and Technology\\
\authorrefmark{6}
Speechocean\\
Corresponding Email:  qyhong@xmu.edu.cn, lilin@xmu.edu.cn
}
}

\maketitle
\thispagestyle{empty}
\renewcommand{\thefootnote}{\alph{footnote}}
\footnotetext[1]{These authors contributed equally to this work.}

\renewcommand{\thefootnote}{\arabic{footnote}}
\begin{abstract}
This paper introduces the sixth Oriental Language Recognition (OLR) 2021 Challenge, which intends to improve the performance of language recognition systems and speech recognition systems within multilingual scenarios. The data profile, four tasks, two baselines, and the evaluation principles are introduced in this paper.
In addition to the Language Identification (LID) tasks, multilingual Automatic Speech Recognition (ASR) tasks are introduced to OLR 2021 Challenge for the first time. The challenge this year focuses on more practical and challenging problems, with four tasks: (1) constrained LID, (2) unconstrained LID, (3) constrained multilingual ASR, (4) unconstrained multilingual ASR. Baselines for LID tasks and multilingual ASR tasks are provided, respectively. The LID baseline system is an extended TDNN x-vector model constructed with Pytorch. A transformer-based end-to-end model is provided as the multilingual ASR baseline system. These recipes will be online published, and available for participants to construct their own LID or ASR systems. The baseline results demonstrate that those tasks are rather challenging and deserve more effort to achieve better performance.
\end{abstract}

\section{Introduction}

The oriental languages, as part of the various languages around the world, can be grouped into several language
families, such as Austroasiatic languages (e.g., Vietnamese,
Cambodia)~\cite{sidwell201114}, Tai-Kadai languages (e.g., Thai, Lao), Hmong-Mien languages (e.g., some dialects in south China), Sino-Tibetan languages (e.g., Chinese Mandarin), Altaic languages (e.g., Korea, Japanese) and Indo-European languages (e.g., Russian)~\cite{ramsey1987languages},~\cite{shibatani1990languages},~\cite{comrie1996russian}.
With the worldwide population movement and communication, more and more multilingual phenomena are emerging in real life.
The oriental languages themselves also influence each other via multilingual interaction, leading to complicated linguistic evolution. 
These complicated multilingual phenomena are worthy of being studied.

Language identification refers to identify the language categories from the given utterance, and it usually presents at the front end of speech processing systems.
Automatic speech recognition is to convert human speech into text.
Existing multilingual speech recognition system can be divided into two categories. One kind of multilingual speech recognition system adopts the cascade architecture, which first determines the language through an LID system and then transcribe the speech using the monolingual ASR system~\cite{2008Language},~\cite{2014An}. The other kind of multilingual speech recognition system uses an end-to-end holistic architecture that collects data from all languages and trains a single model that can handle  languages identification and speech recognition altogether~\cite{2017Language}.
In spite of the brilliant progress, there are still difficult issues decaying the performance of LID and multilingual ASR systems, such as the cross-channel issue, the lack of training resources and the noisy environment.

The oriental language recognition (OLR) challenge is organized annually, aiming at improving the research on multilingual phenomena and advancing the development of multilingual speech technologies.
The challenge has been conducted five times since 2016, namely AP16-OLR~\cite{wang2016ap16}, AP17-OLR~\cite{tang2017ap17}, AP18-OLR~\cite{tang2018ap18}, AP19-OLR~\cite{tang2019ap19} and  AP20-OLR~\cite{2020AP20}, each attracting dozens of teams around the world.



The OLR 2020 Challenge involved more languages, dialects and real-life data, and focused on three challenging tasks:
   (1) cross-channel LID, which was inherited from OLR 2019;
   (2) dialect identification, introduced to OLR for the first time;
   (3) noisy LID, also newly introduced, considering the general importance of noisy speech processing.
In the first task, the champion system, LORIA-Inria-Multispeech, achieved a $C_{avg}$ of 0.0239 and an EER of 2.47\%.
In the second task, the system submitted by Phonexia achieved a $C_{avg}$ of 0.0738 and an EER of 11.97\%.
And in the third task, the system submitted by LORIA-Inria-Multispeech achieved a $C_{avg}$ of 0.0374 and an EER of 4.07\%. 
From these results, one can see that for the cross-channel condition, the task remains challenging.  More details about the past five challenges can be found on the challenge website.\footnote{http://olr.cslt.org}

\begin{table*}[htbp]
      \setlength{\abovecaptionskip}{-0.1cm} 
		\setlength{\belowcaptionskip}{-0.2cm}
		\setlength\tabcolsep{4.5pt}
\centering

  \label{tab:ol10}
\caption{Data Allowed for Constructing Systems }

\begin{tabular}{|c|c|c|c|c|c|c|c|c|c|c|c|c|} 
\hline
\multirow{2}{*}{Language} &
\multirow{2}{*}{Code} & OLR2016       & \multicolumn{3}{c|}{OLR2017} & OLR2018 & \multicolumn{2}{c|}{OLR2019} & \multicolumn{2}{c|}{OLR2020} & \multicolumn{2}{c|}{Total}  \\ 
\cline{3-13}
                 &         & train\&dev(OL7) & train(OL3) & dev  & test     & test    & dev & test                   & train(dailect) & test        & Utterances & Duration       \\ 
\hline
Cantonese & ct-cn                     & 5760          & 0          & 1920 & 2556     & 2558    & 0   & 1800                   & 0              & 3943        & 18537      & 25.23h         \\ 
\hline
Mandarin & zh-cn                     & 5400          & 0          & 1800 & 2400     & 2400    & 500 & 3449                   & 0              & 3310        & 19259      & 25.3h          \\ 
\hline
Indonesian & id-id                     & 5760          & 0          & 1920 & 2557     & 2557    & 0   & 1800                   & 0              & 1800        & 16394      & 21.66h         \\ 
\hline
Japanese & ja-jp                     & 5760          & 0          & 1920 & 2548     & 2544    & 500 & 3424                   & 0              & 3777        & 20473      & 18.99h         \\ 
\hline
Russian & ru-ru                     & 5400          & 0          & 1800 & 1796     & 2394    & 500 & 3441                   & 0              & 3450        & 18781      & 27.44h         \\ 
\hline
Korean & ko-kr                     & 5400          & 0          & 1800 & 2398     & 2399    & 0   & 1800                   & 0              & 3423        & 17220      & 19.81h         \\ 
\hline
Vietnamese & vi-vn                     & 5400          & 0          & 1800 & 2396     & 2400    & 500 & 3422                   & 0              & 1800        & 17718      & 23.91h         \\ 
\hline
Kazakh & Kazak                     & 0             & 2400       & 1800 & 1800     & 1800    & 0   & 1800                   & 0              & 0           & 9600       & 17.9h          \\ 
\hline
Tibetan & Tibet                     & 0             & 9300       & 1800 & 1800     & 1800    & 500 & 3435                   & 0              & 0           & 18635      & 17.9h          \\ 
\hline
Uyghur & Uyghu                     & 0             & 3740       & 1430 & 1800     & 1800    & 500 & 3404                   & 0              & 0           & 12674      & 24.69h         \\ 
\hline
Hokkien & Minnan                    & 0             & 0          & 0    & 0        & 0       & 505 & 0                      & 8000           & 1998        & 10503      & 19.55h         \\ 
\hline
Shanghainese & Shanghai                  & 0             & 0          & 0    & 0        & 0       & 505 & 0                      & 8000           & 1800        & 10305      & 14.98h         \\ 
\hline
Sichuanese & Sichuan                   & 0             & 0          & 0    & 0        & 0       & 505 & 0                      & 8000           & 1800        & 10305      & 13.72h         \\ 
\hline
Thai & th-th                     & 0             & 0          & 0    & 0        & 0       & 0   & 0                      & 0              & 2000        & 2000       & 1.83h          \\ 
\hline
Telugu & te-in                     & 0             & 0          & 0    & 0        & 0       & 0   & 0                      & 0              & 1992        & 1992       & 3.37h          \\ 
\hline
Malay & ms-my                     & 0             & 0          & 0    & 0        & 0       & 0   & 0                      & 0              & 2000        & 2000       & 3.72h          \\ 
\hline
Hindi & hi-in                     & 0             & 0          & 0    & 0        & 0       & 0   & 0                      & 0              & 1952        & 1952       & 3.39h          \\
\hline
\end{tabular}
  \begin{tablenotes}

  \item[a] Male and Female speakers are balanced.
  \item[b] All data in the table except the last column is the number of utterances.
  \item[c] The number of total utterances might be slightly smaller than expected, due to the quality check.
  \end{tablenotes}
\end{table*}

Based on the experience of the last five challenges and the calling from industrial application, we propose the sixth OLR challenge. This new challenge, denoted by OLR 2021 Challenge, involves more languages/dialects and focuses on more practical and challenging tasks.
In this new challenge, we set four tasks, two on LID and two on ASR:
(1) constrained LID, a cross domain identification task with constrained training condition, 
(2) unconstrained LID, a cross domain identification task with unconstrained training condition, and the test utterances in this task are obtained from real-life environments,
(3) constrained multilingual ASR, only the data provided by the organizer can be used to train the acoustic and language system.
(4) unconstrained multilingual ASR, any data is allowed to be used to train the acoustic and language models in addition to the data provided.

In the rest of the paper, we will present the data profile and the evaluation plan of the OLR 2021 challenge. To assist participants to build their own submissions, we provide two baseline systems for LID and multilingual ASR, respectively.

\section{Data Profile}

  Participants of OLR 2021 Challenge can request the following databases for system construction.
  The information of the databases are summarized in Table~\ref{tab:ol10}.
  All these data can be used to train their submission systems.

  \begin{itemize}
  
  \item OLR16-OL7: The standard database for OLR 2016 Challenge, including training set and development set for OLR 2016 Challenge.
  
  \item OLR17-OL3: A database provided by the M2ASR project, involving three new languages.
  
  \item OLR17-dev: The development set for OLR 2017 Challenge. 
  
  \item OLR17-test: The standard test set for OLR 2017 Challenge. 
  
  \item OLR18-test: The standard test set for OLR 2018 Challenge. 
  
  \item OLR19-dev: The standard development set for OLR 2019 Challenge. 
  
  \item OLR19-test: The standard test set for OLR 2019 Challenge.
  
  \item OLR20-dialect: The newly provided training set in OLR 2020 Challenge, including three kinds of Chinese dialects.

  \item OLR20-test: The standard test set for OLR 2020 Challenge.
  \end{itemize}

  Besides the speech signals, the OLR16-OL7, OLR17-OL3, OLR20-dialect and OLR20-test datasets also provide lexica of all the corresponding languages, as well as the transcriptions of all the training utterances. These resources allow training acoustic-based or phonetic-based language recognition systems. And training phone-based speech recognition systems is also possible.
  
  To address the reverse order of Kazakh text, we have recoded the Kazakh transcript so that it is displayed. Table~\ref{tab:transcript} shows some special tags used in the transcriptions.
  
  \begin{table*}[htb]
  \begin{center}
      \setlength{\abovecaptionskip}{0.1cm} 
		\setlength{\belowcaptionskip}{-0.2cm}
  \caption{Special Tags Instructions in the Transcript}
  \label{tab:transcript}
   	\begin{tabular}{|c|c|} 
   		\hline
   		Tags & Definition  \\
   		\hline
   		$**$ & \makecell[l]{Unintelligible speech, words or stretches of speech that are completely unintelligible.} \\
   		\hline
   		$\#$ & \makecell[l]{Filler words, which are “words” that speakers use to indicate hesitation or to maintain control of a conversation \\while thinking of what to say next.}  \\
   		\hline
   	    $<SPK/>$ &  \makecell[l]{Speaker noise: The various sounds and noises made by the speaker that are not part of the prompted text.}\\
   	    \hline
   		$<STA/>$ & \makecell[l]{Stationary noise: This category contains background noise that is not intermittent and has a more or less stable \\amplitude spectrum over some times.} \\
   		\hline
   		$<NON/>$ & \makecell[l]{Non-human noise: This category contains noises of an intermittent nature. These noises typically occur only once \\like a door slam, dropping something or mouse clicking.}  \\
   		\hline
   		$<NPS/>$ & \makecell[l]{Non-Primary Speakers' noise were transcribed as $<NPS/>$.}\\
   		\hline
   	\end{tabular}
   	  \end{center}
  \end{table*}


  \subsection{OLR16-OL7}

  The OLR16-OL7 database was originally created by Speechocean, targeting for various speech processing tasks.
  It was provided as the standard training and development data in AP16-OLR.
  The entire database involves 7 datasets, each in a particular language. The seven languages are:
  Mandarin, Cantonese, Indonesian, Japanese, Russian, Korean and Vietnamese.
  The data volume for each language is about $10$ hours of speech signals recorded in reading style.
  The signals were recorded by mobile phones, with a sampling rate of $16$ kHz  and a sample size of $16$ bits.

  For Mandarin, Cantonese, Vietnamese and Indonesia, the recording was conducted in a quiet environment.
  As for Russian, Korean and Japanese, there are $2$ recording sessions for each speaker: the first session
  was recorded in a quiet environment and the second was recorded in a noisy environment.
  The basic information and the details of the database can be found in the challenge website or the
  description paper~\cite{wang2016ap16}.

  \subsection{OLR17-OL3}

  The OLR17-OL3 database contains 3 languages: Kazakh, Tibetan and Uyghur, all are minority languages in China.
  This database is part of the Multilingual Minorlingual Automatic Speech Recognition (M2ASR) project, which is supported by the National Natural Science Foundation of China (NSFC). The project is a three-party collaboration, including Tsinghua University, the Northwest National University, and Xinjiang University~\cite{wangm2asr}. The aim of this project is to construct speech recognition systems for five minor languages in China (Kazakh, Kirgiz, Mongolia, Tibetan and Uyghur). 
  More information about this database, please refer to the web site of the project.\footnote{http://m2asr.cslt.org}

  \subsection{OLR17-test}

  The OLR17-test database is a dataset provided by Speechocean. This dataset contains 7 languages as in OLR16-OL7, each containing $1800$ utterances. The recording conditions are the same as OLR16-OL7. This database is used as part of the test set for the OLR 2017 Challenge.
  The sentences of each language in OLR17-OL3-test are randomly selected from the original M2ASR corpus.
  The data volume for each language in OLR17-OL3 is about $10$ hours of speech signals recorded in reading style.
  The signals were recorded by mobile phones, with a sampling rate of $16$ kHz and a sample size of $16$ bits.
  We selected $1800$ utterances for each language as the development set (OLR17-OL3-dev), and the rest was used as the
  training set (OLR17-OL3-train). The test set of each language involves $1800$ utterances, and was provided separately and denoted by OLR17-OL3-test.
  Compared to OLR16-OL7, OLR17-OL3 contains much more variations in terms of recording conditions and
  the number of speakers, which may inevitably  increase the difficulty of the challenge task.

  \subsection{OLR18-test}
  The OLR18-test database is the standard test set for OLR 2018 Challenge.
  Like the OLR17-test database,
  OLR18-test contains the same target $7$ languages, each containing $1800$ utterances, while OLR18-test also contains utterances from several interference languages.The recording conditions are the same as OLR17-test.

  \subsection{OLR19-test}
  The OLR19-test database is the standard test set for OLR 2019 Challenge,
  which includes 3 parts corresponding to the 3 LID tasks respectively, precisely OLR19-short, OLR19-channel and OLR19-zero.

  \subsection{OLR20-dialect}
  OLR20-dialect is the training set provided by Speechocean. It includes three kinds of Chinese dialects, namely Hokkien, Sichuanese and Shanghainese. The utterances of each language are about 8000. The signals were
  recorded by mobile phones, with a sampling rate of $16$ kHz  and a sample size of $16$ bits.

  \subsection{OLR20-test}
  The OLR20-OLR-test database is the standard test set for OLR 2020 Challenge, which includes 3 parts corresponding to the 3 LID tasks respectively, precisely OLR20-OLR-channel-test, OLR20-OLR-dialect-test and OLR20-OLR-noisy-test.
  
  \subsection{OLR21-test}
   For the OLR 2021 Challenge, the trials of the four tasks will be divided into two subsets: a progress subset, and a test subset. The progress subset will comprise 30\% of the trials and will be used to monitor progress in the leaderboard. The remaining 70\% of the trials will form the test subset, and will be used to generate the official results which are the base of the final ranking. 
   The OLR21-test database is the standard test set for the OLR 2021 challenge, which contains two parts: OLR21-cross-domain-test and OLR21-wild-test.

\begin{itemize}
 
  \item OLR21-cross-domain-test: This subset is designed for three tasks: the constrained LID task, the constrained multilingual ASR task, and the unconstrained multilingual ASR task. It contains $13$ languages, and was recorded by different recording equipments in different environments. The $13$ languages are Indonesian, Japanese, Russian, Korean, Vietnamese, Mandarin, Cantonese (China), Sichuanese (China), Shanghainese (China), Hokkien (China), Tibetan (China), Kazakh (China), and Uyghur (China).

  \item OLR21-wild-test: This subset is designed for the unconstrained LID task, which contains $17$ languages: Indonesian, Japanese, Russian, Korean, Vietnamese, Thai, Malay, Telugu, Hindi, English (British and American), Kazakh (China), Tibetan (China), Uyghur (China), Mandarin, Sichuan (China), Shanghainese (China), Hokkien (China). Utterances in this subset are obtained from real-life environments.

  \end{itemize}

  \section{OLR 2021 Challenge}
  
 Following the last five challenges, the LID task in OLR 2021 (OLR-LID) is defined as follows: Given a segment of speech and  a language hypothesis (i.e., a target language of interest to be detected), the task is to decide whether that target language was in  fact  spoken  in  the given segment (yes or no), based on an automated analysis of the data contained in the segment.
 In addition, we add multilingual ASR tasks (OLR-ASR) to further advance speech technology research in multilingual environments. The multilingual ASR task is defined as: The system needs to output the corresponding transcripts for the given utterances in unknown language. Besides, there is only one language in each utterance.

  \subsection{OLR-LID}

  The OLR-LID challenge includes two tasks as follows:

  \begin{itemize}
  \item Task 1: Constrained Task (Cross Domain)
  
    This task is a close-set identification task, which means the language of each utterance is among the known 13 target languages (Indonesian, Japanese,  Russian, Korean, Vietnamese, Mandarin, Cantonese, Sichuanese, Shanghainese, Hokkien, Tibetan, Kazakh and Uyghur), but utterances were recorded in different environments. And only the data provided by the organizer can be used to build the LID system.

  \item Task 2: Unconstrained Task (Wild Data)

    This task is also a close-set identification task, but test data from wild, which means utterances are obtained from real-life environments. It is therefore more challenging than the constrained task. In this task, any data (except the evaluation data) you can access is allowed to build the system. This task involved 17 languages (Indonesian, Japanese, Russian, Korean, Vietnamese, Thai, Malay, Telugu, Hindi, English, Kazakh, Tibetan, Uyghur, Mandarin, Sichuan, Shanghainese, Hokkien). Utterances in this subset are obtained from real-life environments.



  \end{itemize}

  \subsubsection{System input/output}

  The input to the LID system is a set of speech segments in unknown languages.
  For task 1, those speech segments are within the $13$ known target languages.
  For task 2, speech segments are within the $17$ known target languages.
  The task of the LID system is to determine the confidence that a language is contained in a speech segment. More specifically,
  for each speech segment, the LID system outputs a score vector $<\ell_1, \ell_2, ..., \ell_{10}>$,
  where $\ell_i$ represents the confidence that language $i$ is spoken in the speech segment.  
   The scores should be comparable across languages and segments.
  This is consistent with the principles of LRE15~\cite{lre15}, but differs from that of LRE09~\cite{lre09} where an explicit decision is required for each trial.

  In summary, the output of an OLR submission will be a text file, where each line contains a speech segment plus a score vector for this segment, e.g.,

  \vspace{0.5cm}
  \begin{tabular}{ccccccccc}
          & lang$_1$   & lang$_2$   & ... & lang$_9$  & lang$_{10}$\\
  seg$_1$ & 0.5  & -0.2 &  ...& -0.3 & 0.1    \\
  seg$_2$ & -0.1 & -0.3 &  ...& 0.5 & 0.3    \\
  ...   &      &     &  ... &      &
  \end{tabular}

  \subsubsection{Training condition}

The training condition is defined as the amount of resources used to build a LID system.

    \begin{itemize}
  \item For task 1, the use of additional training materials is forbidden, with the exception of non-speech data.
  The audio resources allowed to use are:
        OLR16-OL7, OLR17-OL3, OLR17-test, OLR17-dev, OLR18-test, OLR19-test, OLR19-dev, OLR20-dialect, OLR20-test.
        
 \item For task 2, any publicly available or proprietary data is allowed for system training and development. Participating teams must provide a detailed description of the speech and non-speech data resources as well as any pre-trained models used during the training and development of their systems.
  \end{itemize}

  \subsubsection{Test condition}


  \begin{itemize}
  \item All the trials should be processed. Scores of lost trials will be interpreted as -$\inf$.
  \item The speech segments in each task should be processed independently, and each test segment in a group should be processed independently too. Knowledge from other test segments is not allowed to use (e.g., score distribution of all the test segments).
  \item Information of speakers is not allowed to use.
  \item Listening to any speech segments is not allowed.
  \end{itemize}

  \subsubsection{Evaluation metrics}

  As in LRE15, the OLR 2021 Challenge chooses $C_{avg}$ as the principle evaluation metric.
  First define the pair-wise loss that composes the missing and
  false alarm probabilities for a particular target/non-target language pair:

  \[
  C(L_t, L_n)=P_{Target} P_{Miss}(L_t) + (1-P_{Target}) P_{FA}(L_t, L_n)
 \]

  \noindent where $L_t$ and $L_n$ are the target and non-target languages, respectively; $P_{Miss}$ and
  $P_{FA}$ are the missing and false alarm probabilities, respectively. $P_{Target}$ is the prior
  probability for the target language, which is set to $0.5$ in the evaluation. Then the principle metric
  $C_{avg}$ is defined as the average of the above pair-wise performance:


  \[
   C_{avg} = \frac{1}{N} \sum_{L_t} \left\{
  \begin{aligned}
    & \ P_{Target} \cdot P_{Miss}(L_t) \\
    &  + \sum_{L_n}\ P_{Non-Target} \cdot P_{FA}(L_t, L_n)\
  \end{aligned}
  \right\}
  \]

  \noindent where $N$ is the number of languages, and $P_{Non-Target}$ = $(1-P_{Target}) / (N -1 )$.
  We have provided the evaluation scripts for system development.

 \subsection{OLR-ASR}

  We set up two tasks for participants to study multilingual speech recognition with different limits on the use of training data. The OLR2021-ASR includes two tasks as follows:

  \begin{itemize}
  \item Task 1: Constrained Task 
  
This task is a data resources constrained task, only the data provided by the organizer can be used.

\item Task 2: Unconstrained Task

This task is a task with unconstrained data resources and any data can be used for training and optimization. 

  \end{itemize}

  \subsubsection{System input/output}

  The input to the ASR system is a set of speech utterances in unknown languages. Both two ASR tasks and LID task 1 share the same evaluation set, where the speech utterances are within the $13$ known target languages. For each speech utterance, the ASR system should output the corresponding transcript. In summary, the output of an ASR task submission will be a text file, within which each line contains, e.g.,

  \vspace{0.5cm}
  \begin{center}

  \begin{tabular}{ccccccccc}
  HOW ARE YOU & (utt$_1$)   \\
  CAN YOU COME HERE & (utt$_2$) \\
  ...   &  ...
  \end{tabular}
  \end{center}
  \subsubsection{Training condition}
The OLR2021-ASR offers constrained task and unconstrained task. Referring to OpenASR20 Challenge\footnote{https://www.nist.gov/itl/iad/mig/openasr-challenge}, Table~\ref{tab:condition} shows the details about permissible additional data resources tasks for the two multilingual ASR.
    \begin{itemize}
  \item For task 1, only the speech audio data provided in this project can be used for training.
        
  \item For task 2,  any publicly labeled audio data and any unlabeled audio data are available to further improve system performance, but teams must provide the source of the data in the final system description submission.
  
  For both tasks, additional text data is permissible for training in constrained task. Any such additional text training data must be specified in sufficient detail in the system description.

       \end{itemize}
       

  \begin{table}[htbp]
  \caption{Rules for Using Data outside the Set}
  \centering
  \label{tab:condition}
 \resizebox{\linewidth}{9mm}{
 \setlength\tabcolsep{3pt}
   	\begin{tabular}{|lll|} 
   		\hline
   		Resources beyond the data provided in this challenge & task1 & task2  \\
   		\hline
   		Speech data & no &  yes \\
   		Text data & yes & yes  \\
   		Pretrained models trained on speech data & no &  yes \\
   		Pretrained models trained on text data & yes &  yes \\
   		TTS audio data trained with data provided in this project & yes & yes  \\
   		Non-speech acoustic data (noise etc.) &  yes &  yes \\
   		\hline
   	\end{tabular}}
  \end{table}

  \subsubsection{Test condition}

  \begin{itemize}
  \item The speech segments in each task should be processed independently, and each test segment in a group should be processed independently too. Knowledge from other test segments is not allowed to use (e.g., score distribution of all the test segments).
  \item Listening to any speech segments is not allowed.
  \end{itemize}

  \subsubsection{Evaluation metrics}

Considering that the dataset consists of languages with different grammatical rules, we use Character Error Rate (CER) as the primary metric computed on the submitted output. CER is computed as the sum of deletion, insertion, and substitution errors in the ASR output compared to the reference transcription, divided by the total number of characters in the reference transcription:
$$
CER=\frac{\#Deletions+\#Insertions+\#Substitutions}{\#ReferenceWords}
$$
We use the CERs of the entire evaluation set as ranking metrics and provide separate CERs for each language as a reference. Punctuation marks and special labels are not included in the calculation of CER.

  \section{Baseline systems}

Two baseline systems are provided in OLR challenge 2021, the baseline LID systems for OLR-LID and the baseline ASR systems for OLR-ASR. All the experiments are conducted with Pytorch~\cite{tong2021asv}. The purpose of these experiments is to present a reference for the participants, rather than a competitive submission. The recipes can be downloaded from the website of the challenge.\footnote{http://cslt.riit.tsinghua.edu.cn/mediawiki/index.php/OLR\_Challenge\_2021}

\subsection{The LID Baseline System}

The baseline LID systems in this challenge was an extended TDNN x-vector model~\cite{snyder2018x},~\cite{snyder2018spoken},~\cite{villalba2016etdnn} constructed with Pytorch. The feature extracting and back-ends were all conducted with Kaldi. 

  \subsubsection{The extended TDNN x-vector system}

  Compared to the traditional x-vector, the extended TDNN x-vector structure used a slightly wider temporal context in the TDNN layers and interleave dense layers between TDNN layers, which leaded to a deeper x-vector model. This deep structure was trained to classify the $N$ languages in the training data with the cross entropy (CE) loss. After training, embeddings called `x-vector' were extracted from the affine component of the penultimate layer.

  We trained the baseline systems with all the data showed in Table~\ref{tab:ol10}.
  Before training, we adopted speed perturbation for data augmentation, to increase the amount and diversity of the training data. We applied a speed factor of 0.9 or 1.1 to slow down or speed up the original recording. Finally, the augmented copies of the original recording were added to the original data set to obtain a 3-fold combined training set.

  The acoustic features involved 20-dimensional Mel frequency cepstral coefficients (MFCCs) with the 3-dimensional pitch, and the energy VAD was used to filter out nonspeech frames.

  
  Linear discriminative analysis (LDA) trained on the enrollment set was employed to promote language-related information when the embeddings was extracted from the model. The dimensionality of the LDA projection space was set to 100. After the LDA projection and centering, the logistic regression (LR) trained on the enrollment set was used to compute the score of a trial on a particular language.


  \subsubsection{Performance results}

  The primary evaluation metric in OLR 2021 Challenge  is $C_{avg}$. Besides that, we also present the performance in terms of equal error rate (EER). These metrics evaluate system performance from different perspectives, offering a whole picture of the capability of the tested system.  The performance of baselines is evaluated on the progress subset of OLR-2021-test.
  
 Table~\ref{tab:11} show the utterance-level C$_{avg}$ and EER results on the progress subset.
  
\begin{table}[h]
      \setlength{\abovecaptionskip}{-0.1cm} 
		\setlength{\belowcaptionskip}{-0.2cm}
\centering
\caption{  $C_{avg}$ and EER results on the progress subset}
\label{tab:11}
\begin{tabular}{|c|c|c|} 
\hline
Dataset                &  $C_{avg}$   & EER    \\ 
\hline
progress subset & 0.0826 & 9.038\%  \\ 
\hline
\end{tabular}
\end{table}



  \subsection{The ASR Baseline System}

 \begin{table*}[htbp]
       \setlength{\abovecaptionskip}{-0.1cm} 
		\setlength{\belowcaptionskip}{-0.2cm}
		\setlength\tabcolsep{5pt}
 \centering
 \caption{CER results on the progress subset}
 \label{CER}
 \begin{tabular}{|c|c|ccccccccccccc|} 
  \hline
Dataset   & Total & zh-cn & Minnan & Shanghai & Sichuan & ct-cn & id-id & ja-jp & ko-kr & ru-ru & vi-vn & Kazak & Tibet & Uyghu   \\ 
  \hline
 progress subset & 39.4\% & 116.8\% & 69.3\% & 35.9\% & 34.5\% & 47.0\% & 9.2\% & 67.3\% & 34.2\% & 35.5\% & 31.1\% & 35.1\% & 52.7\% & 21.0\%  \\ 
  \hline
 \end{tabular}
\end{table*}

For the multilingual ASR tasks, we trained the baseline with a combined dataset including OLR16-OL7, OLR17-OL3, OLR20-dialect, OLR20-test. The number of languages in these training sets can cover the $13$ languages in the test set, while supplementing the training dataset with some special channels. 
Before training, we adopted the data augmentation, including speed perturbation and SpecAugment~\cite{2019SpecAugment}, to increase the amount and diversity of the training data. For speed perturbation, we applied a speed factor of 0.9 or 1.1 to slow down or speed up the original recording. Finally, two augmented copies of the original recording were added to the original data set to obtain a 3-fold combined training set. We used SpecAugment method to direct process the spectrogram which performed time wraping, frequence masking and time masking for the spectrogram features. Besides, we didn't train additional language models.

  \subsubsection{Transformer system}

    For the ASR tasks, we used a transformer-based end-to-end model as a baseline, which combined encoder-decoder structure with an attention mechanism. End-to-end systems generally fold the acoustic, lexicon, and language models into a single network. They save the effort on language-specific processing, making it easier to apply them to the multilingual ASR task. Therefore, it had been used in multilingual speech recognition~\cite{2018Multilingual}. Transformer was firstly introduced in neural machine translation and achieved superior performance recently in speech recognition~\cite{2018Speech},~\cite{2019The}. The encoder can transfer the input feature sequence to a high-level representation sequence. The decoder decodes the transferred sequence in the way of autoregressive and get the prediction sequence step by step. The encoder and decoder of the transformer are composed of stacked self-attention and position-wise feed-forward layers. In the transformer model multi-head attention plays a significant role, which can obtain information from different representation subspace and each head could focus on a different subspace. In order to enable the model to make rational use of the location information of the sequence, the model accepts positional codes to obtain the location information of each input element.
    
    Besides, we adopted a language-independent architecture so that all the target languages could share the same network architecture and parameters. This architecture required the output vocabulary includes characters of all the languages. This setting made it possible to train a single network for all languages in a language-independent manner.

    We used ESPnet~\cite{2018ESPnet} toolkit to build our baseline. For audio data, 80-dimensional mel-filterbanks features with 3-dimensional pitch features were used as the input feature. We used the standard configuration of ESPnet transformer which contained 12-layer encoder and 6-layer decoder with 2048-dimensional each layer. The attention sub-layer was 256-dimensional and used 4 attention heads. The whole network was trained for 30 epochs and warm-up was used for the first 25000 iterations. We used the characters from the training set as the output units of the model.

   \subsubsection{Performance results}

   The primary evaluation metric in OLR2021 ASR task is the CER on entire evaluation set. Besides that, we also present separate CERs for each language. Participants are able to analyze the performance of the system in total and in individual languages. 
   Participants are encouraged to design their own development sets, so that we do not provide the referenced development set results. Table~\ref{CER} shows the CER results on the progress subset.
   
   The transformer baseline achieves the CER with 39.4\% in the progress subset. It is worth noting that we have analysed the decoding results and found that the baseline system is poor at recognising noisy data and cross-channel data, resulting in high CERs in some languages. Teams may need to tune for such issues.

  \section{Conclusions}

  In this paper, we present the data profile, four task definitions and two baselines of the OLR 2021 Challenge. In this challenge, besides the presented data sets in past challenges, test sets of OLR 2021 Challenge are provided for participants, and more languages are included in the test sets. The OLR challenge 2021 are with four tasks: (1) constrained LID, (2) unconstrained LID, (3) constrained multilingual ASR, (4) unconstrained multilingual ASR. Baseline systems of LID and ASR are released to assist participants to construct a reasonable starting system. Given the results on baseline systems, these four tasks defined by OLR 2021 challenge are rather challenging and are worthy of careful study.
 All the data resources are free for the participants, and the recipes of the baseline systems can be freely downloaded.


  \section*{Acknowledgments}

  This work was supported by the National Natural Science Foundation of China No.61876160, No.61633013 and No.62001405.

  We would like to thank Ming Li at Duke Kunshan University, Xiaolei Zhang at Northwestern Polytechnical University, Miao Zhao and Zhiyuan Tang for their help in organizing this OLR 2021 Challenge.

  \bibliographystyle{IEEEtran}
  \bibliography{olr}

  \end{document}